\documentclass[letterpaper, 10 pt, conference]{ieeeconf}  % Comment this line out if you need a4paper

\IEEEoverridecommandlockouts                              % This command is only needed if 
\usepackage{amsmath}
\usepackage{amssymb}
\usepackage{graphicx}
\usepackage[Symbol]{upgreek}
\usepackage{epstopdf}
\usepackage{subcaption}
\usepackage{enumerate}
\usepackage{paralist}
\usepackage{comment}

\usepackage{float}

\usepackage{latexsym}
\usepackage{multicol}
\usepackage{multirow}
\usepackage{lipsum}

\usepackage{cite}

\usepackage{makecell}
\usepackage{breqn}

\usepackage{verbatim}

\usepackage{color}
\usepackage[dvipsnames]{xcolor}
\usepackage{bm}
\usepackage[font=small,labelfont=bf]{caption}

\usepackage{algpseudocode}
\usepackage{algorithm2e}
\newlength\mylen
\newcommand\myinput[1]{%
  \settowidth\mylen{\KwIn{}}%
  \setlength\hangindent{\mylen}%
  #1\\}

\usepackage[normalem]{ulem}
\usepackage{url}

\DeclareUrlCommand\ULurl{%
  \renewcommand\UrlLeft{\uline\bgroup}%
  \renewcommand\UrlRight{\egroup}}

\newcommand{\vect}[1]{\bm{#1}}

\newcommand{\bvect}[1]{\bar{\vect{#1}}}
\newcommand{\hvect}[1]{\hat{\vect{#1}}}
\newcommand{\dvect}[1]{\dot{\vect{#1}}}
\newcommand{\ddvect}[1]{\ddot{\vect{#1}}}

\newcommand{\vphi}[0]{\vect{\phi}}

\newcommand{\phVar}[0]{s}

\usepackage{courier}
\usepackage{tikz}
\usetikzlibrary{calc}
\def\particulartemplate#1{
  \begin{tikzpicture}[overlay, remember picture]
    \draw let \p1 = (current page.west), \p2 = (current page.east) in
      node[minimum width=\x2-\x1, minimum height=0.1cm, rectangle, fill=yellow!35!white, anchor=north west, align=center, text width=\x2-\x1] at ($(current page.north west) + (0,-0.3)$) {\large \textbf{\texttt{#1}} };
  \end{tikzpicture}
}

\title{\LARGE \bf
From RGB images to Dynamic Movement Primitives for planar tasks
}

\author{Antonis Sidiropoulos and Zoe Doulgeri% <-this % stops a space
\thanks{Authors are with the Automation \& Robotics Lab, Dept. of Electrical
\& Computer Engineering, Aristotle University of Thessaloniki, Greece. 
        {\tt\small \{antosidi, doulgeri\}@ece.auth.gr}}%
\thanks{The research leading to these results has received funding from the European Community’s Framework Programme Horizon 2020 under grant agreement No 871704, project BACCHUS.}%
}

\begin{document}

\maketitle
\thispagestyle{empty}
\pagestyle{empty}

\particulartemplate{
Post-Print version (final draft post-refereeing). \\
Publisher’s version at https://doi.org/10.1109/Humanoids57100.2023.10375217
}

%%%%%%%%%%%%%%%%%%%%%%%%%%%%%%%%%%%%%%%%%%%%%%%%%%%%%%%%%%%%%%%%%%%%%%%%%%%%%%%%
\begin{abstract}
    DMP have been extensively applied in various robotic tasks thanks to their generalization and robustness properties. However, the successful execution of a given task may necessitate the use of different motion patterns that take into account not only the initial and target position but also features relating to the overall structure and layout of the scene. To make DMP applicable to a wider range of tasks and further automate their use, we design a framework combining deep residual networks with DMP, that can encapsulate different motion patterns of a planar task, provided through human demonstrations on the RGB image plane. We can then automatically infer from new raw RGB visual input the appropriate DMP parameters, i.e. the weights that determine the motion pattern and the initial/target positions. We compare our method against another SoA method for inferring DMP from images and carry out experimental validations in two different planar tasks.
    % We experimentally validate our method in the task of unveiling the stem of a grape-bunch from occluding leaves using on a mock-up vine setup and compare it to another SoA method for inferring DMP from images. 
\end{abstract}

%%%%%%%%%%%%%%%%%%%%%%%%%%%%%%%%%%%%%%%%%%%%%%%%%%%%%%%%%%%%%%%%%%%%
%%%%%%%%%%%%%%%%%%%%%%%%%%%%%%%%%%%%%%%%%%%%%%%%%%%%%%%%%%%%%%%%%%%%

\section{Introduction} \label{sec:Introduction}

The objective of employing robots in everyday applications and tasks has been the inspiration and driving force of a lot of research over the last years.
Learning from demonstration has emerged as a promising and intuitive approach for transferring human skills to robots, endowing robots with greater versatility and flexibility and making them more predictable and acceptable by humans \cite{PbD_Billard, Schaal_PbD_1999}.
Various models have been proposed for the purpose of encoding and generalizing human demonstrations of a given task. Such a model is the Dynamic Movement Primitives (DMP) \cite{Ijspeert2013}, which have been utilized in a plethora of robotic applications. DMP are widely acknowledged thanks to their ease of training, generalization and robustness properties. Nevertheless, a DMP can only learn a single stereotypical motion pattern, which may be inadequate in tasks where the structure and/or the layout of the scene calls for different motion patterns. Moreover, the generalization is based only on the initial and target position, which must further be provided explicitly, e.g. from external perception mechanisms.
% They lack the ability to generalize based on rich input context, such as entire RGB images, or produce different trajectory patterns.

% DMP with a Fuzzy Gaussian Mixture Model was proposed in \cite{DMP_FGMM_2019} for encoding multiple demonstrations. However, the average statistics are learned this way, which may be uninformative if the demonstrations vary a lot, i.e. they represent different motion patterns.

Encoding multiple motion patterns from demonstrations with DMP can be addressed using a motion primitive library \cite{DMP_Pastor_2009}, where a separate DMP is trained for each different demonstration. However, generalization to new conditions requires the selection of the appropriate DMP, which is far from straightforward. 
Some works use the target and/or the initial position to infer the DMP weights that will be used for generalization. In particular, Locally Weight Regression was employed in \cite{Ude_task_specific_generalization_2010} to  infer the DMP weights given a desired new target. 
In \cite{AE_DMP_2015}, Deep Auto-encoders are used to reduce the dimensionality of the input data and train the DMP on a latent space. This idea was extended in \cite{Var_AE_DMP_2016}, enabling also the encoding of multiple demonstrations, where the generalization was based on the initial and target position. 

Considering only the initial and target position may be insufficient in cases where additional features in the scene may affect the generalization, like an object in the scene or an intermediate position.
To this end, the authors in \cite{TP_DMP_2018} proposed to use Gaussian Mixture Models to infer the DMP parameters from task specific features of the scene,
% termed in \cite{TP_DMP_2018} as ''task related parameters'', 
which in their case involved the initial and target position and the $2$D position of an object that had to be swept into a dustpan. Still the position of the object had to be provided explicitly, so the authors extended their work in \cite{DMP_CNN_2017} by employing a Convolution Neural Network (CNN) to extract the position of the object from an RGB image of the scene and feed it along with the target position to a neural network that produces a DMP generated trajectory. 
Both in \cite{TP_DMP_2018} and \cite{DMP_CNN_2017} the initial and target positions of the DMP were provided explicitly.
% and the number of the task parameters must be prespecified.

% while the network design in \cite{DMP_CNN_2017} is designed to output only a single task parameter (the $x, y$ coordinates of the object to be swept), without an obvious way to also incorporate additional task parameters.

While all aforementioned works enable the encoding of multiple trajectories with DMP, they only account for the initial and target position or some prespecified task related features, like an intermediate position, to infer the appropriate motion pattern.  
In contrast, inferring DMP directly from raw image inputs enables the exploitation of diverse and rich features like the specific structure of the scene, the layout and geometry of relevant to the task objects etc.
In this direction, the use of deep Neural Networks that take as input a raw image and output directly the DMP parameters has been explored in a few recent publications \cite{DMP_NN_Ude_2020, Arc_DMP_NN_Ude_2021, DNP_2020_Bahl}. In 
\cite{DMP_NN_Ude_2020, Arc_DMP_NN_Ude_2021} a CNN is proposed that given the grayscale image of a depicted digit, it infers the parameters of a DMP, that is used to generate the trajectory for writing the depicted digit. 
% This method was also specialized in \cite{Arc_DMP_NN_Ude_2021} to make it independent of the trajectory's time duration. 
A framework similar to \cite{DMP_NN_Ude_2020} is proposed in \cite{DNP_2020_Bahl}, with the main difference being the use of Reinforcement Learning to learn continuous trajectory policies
% , instead of primitive actions, that are
modeled as DMP. 
% As shown in \cite{DNP_2020_Bahl}, the use of DMP instead of simple action primitives produces smoother trajectories and also achieves higher. 
However, the proposed method is only tested in simulations, without addressing the sim-to-real problem.

In this work, we propose to infer the DMP parameters (initial/target position and motion pattern encoded in the DMP weights) for a planar path/trajectory given the knowledge of the task plane directly from raw RGB images from multiple human demonstrations of the task. 
% The inferred DMP parameters involve not only the initial and target position but also the motion pattern i.e. the DMP weights, without requiring any additional input. 
The considered planar tasks, which are experimentally validated, involve interaction with the environment and hence are affected by the environmental dynamics. In particular, we consider the tasks of pushing aside hinged flexible objects (leaves) to unveil an occluded region of interest (grapebunch stem), and grasping a target object in clutter by pushing aside other objects on a supported surface.
% Such tasks may be for example, pushing of objects lying on a supported surface or pushing aside hinged flexible objects.
% like the leaves of a plant to unveil an occluded crop. 
% In this respect, they are differentiated from  \cite{DMP_NN_Ude_2020, Arc_DMP_NN_Ude_2021}, since  the task in these works involves essentially  the reproduction of a depicted path. Also, in contrast to \cite{DNP_2020_Bahl}, we test our method in real experimental setups and employ the PbD paradigm, that is learning from actual human demonstrations.
To achieve our aim, we use Deep Residual Convolution Networks which have been successfully applied in image recognition and segmentation tasks \cite{Image_recognition_resnet, FCN_seg_2015}. Moreover, we employ the DMP formulation from \cite{Antosidi_Rev_DMP}, which simplifies learning-training.
% , as it avoids additional calculations in order to derive the gradients for back-propagation.
Experimental comparisons with \cite{Arc_DMP_NN_Ude_2021} shows that our proposed method leads to more accurate learning and better generalization in novel scenes, which can be attributed to the more sophisticated network architecture that we employ.

The rest of this paper is organised as follows:
Preliminaries on DMP are provide in Section \ref{sec:DMP_prelim}. The proposed method is presented in Section \ref{sec:Resnet_DMP_architecture}. 
Details on the experimental setup and validation are provided in Section \ref{sec:Experiments}.
Finally, the conclusions are drawn in Section \ref{sec:Conclusions}.
The source code 
% for all conducted simulations and experiments 
is available at {\small \ULurl{github.com/Slifer64/resnet_dmp.git}}

% All works so far work on $2$D. We retain this assumption here too.
% The incremental contribution compare to the aforementioned existing works can be compactly summarize in the following:
% \begin{itemize}
%     \item The structure of the Deep DMP allows it to retain all properties of DMP and also extract meaningful features even for complex tasks due to its deep architecture.
%     \item The task in which our method is tested is more challenging than the previous ones due to the dynamics of the environment, which make the successful execution of the policy more challenging. E.g. some leaves may be partially occluded from other leaves, they are less distinguishable than the objects used in \cite{}, and they are attracted to their initial position, so while being pushed, some leaves may fall back again and hide the stem.
% \end{itemize}

%%%%%%%%%%%%%%%%%%%%%%%%%%%%%%%%%%%%%%%%%%%%%%%%%%%%%%%%%%%%%%%%%%%%
%%%%%%%%%%%%%%%%%%%%%%%%%%%%%%%%%%%%%%%%%%%%%%%%%%%%%%%%%%%%%%%%%%%%

\section{DMP Preliminaries} \label{sec:DMP_prelim}

Here we briefly introduce the basics of the DMP formulation from \cite{Antosidi_Rev_DMP} for $n$-DoFs.
The DMP encodes a demonstrated trajectory $\vect{y}_r(t)$ and can generalize this trajectory starting from a new initial position $\vect{y}_0$ towards a new target/final position $\vect{g}$ with time duration $T_f$. The DMP's evolution is driven by the canonical system, which provides the phase variable $\phVar$ (time substitute), to avoid direct time dependency. The DMP model is given by: 
% and the canonical system are given by:
\begin{align}
    \ddvect{y} &= \ddvect{y}_{\phVar} - \vect{D}(\dvect{y}-\dvect{y}_{\phVar}) - \vect{K}(\vect{y} - \vect{y}_{\phVar} ) \label{eq:GMP_tf_sys} \\
    \dot{\phVar} &= 1/\tau \ , \ \phVar(0) = 0 \ , \ \phVar(T_f) = 1 \label{eq:GMP_x_dot}
\end{align} 
where $\vect{y}$ is the position, $\vect{K}, \vect{D}$ are positive definite matrices and $\vect{y}_{\phVar}$ provides the generalized trajectory reference:
\begin{align} 
    \vect{y}_{\phVar} \triangleq \vect{K}_s(\vect{W}\vphi(\phVar) - \hvect{y}_{0}) + \vect{y}_0 \label{eq:GMP_y_x}
    % &\dvect{y}_{\phVar} = \vect{K}_s \vect{W} \vphi_1(\phVar) \label{eq:GMP_y_x_dot} \\
    % &\ddvect{y}_{\phVar} = \vect{K}_s \vect{W} \vphi_2(\phVar) \label{eq:GMP_y_x_ddot}
\end{align}
where $\vphi(\phVar)^T = [\phi_1(\phVar) \ \cdots \ \phi_K(\phVar)]/{\sum_{k=1}^{K} \phi_k(\phVar)}$, with $\phi_k(\phVar) = e^{-h_k(\phVar-c_k)^2}$ are Gaussian kernels, while the weights $\vect{W} \in \mathbb{R}^{n \times K}$ are optimized using Least Squares (LS) or Locally Weighted Regression (LWR) \cite{Ijspeert2013} so that $\vect{W}\vphi(\phVar) \approx \vect{y}_r$. 
This DMP formulation was shown in \cite{Antosidi_Rev_DMP} to be mathematically equivalent to the classical DMP \cite{Ijspeert2013}, having the same stability and generalization properties and exhibiting the same dynamic behaviour.
Notice that with this DMP formulation, training requires only data from the demonstrated position trajectory as opposed to the classical formulation where data from the velocity and acceleration trajectories are further needed to learn the DMP weights. 
The centers $c_i$ of the Gaussian kernels are equally spaced in time and the inverse widths of the Gaussians are $h_i = \frac{1}{(a_h(c_{i+1}-c_i))^2}$, $i=1,\cdots,N-1$, $h_N = h_{N-1}$, where $a_h > 0$ controls the overlapping between the kernels.
Generalization to a new target $\vect{g}$ is achieved through the spatial scaling matrix $\vect{K}_s \triangleq \text{diag}((\vect{g}-\vect{y}_0) ./ (\hvect{g} - \hvect{y}_{0}))$ 
where $\hvect{g} = \vect{W}\vphi(1)$, $\hvect{y}_{0} = \vect{W} \vphi(0)$ and $./$ stands for the element-wise division.
The reference velocity $\dvect{y}_{\phVar}$ and acceleration $\ddvect{y}_{\phVar}$ can be obtained analytically from \eqref{eq:GMP_y_x} as shown in \cite{Antosidi_Rev_DMP}.
Regarding the canonical system \eqref{eq:GMP_x_dot}, $\tau > 0$ controls the speed of the phase variable's evolution. Choosing a constant $\tau = T_f$, result in the motion duration being approximately equal to $T_f$.

%%%%%%%%%%%%%%%%%%%%%%%%%%%%%%%%%%%%%%%%%%%%%%%%%%%%%%%%%%%%%%%%%%%%
%%%%%%%%%%%%%%%%%%%%%%%%%%%%%%%%%%%%%%%%%%%%%%%%%%%%%%%%%%%%%%%%%%%%

% \cc{
% - Normalize $\rightarrow$ scale in [0 1]

% - Projection $2D$ to $3D$: explain more... Notation on image.

% - Definition of task plane??

% - Define successful unveiling, e.g. $40\%$

% - Report success rate on table
% }

\section{Proposed Method} \label{sec:Resnet_DMP_architecture}

\begin{figure*}[!h]
	\centering
	\includegraphics[scale=0.29]{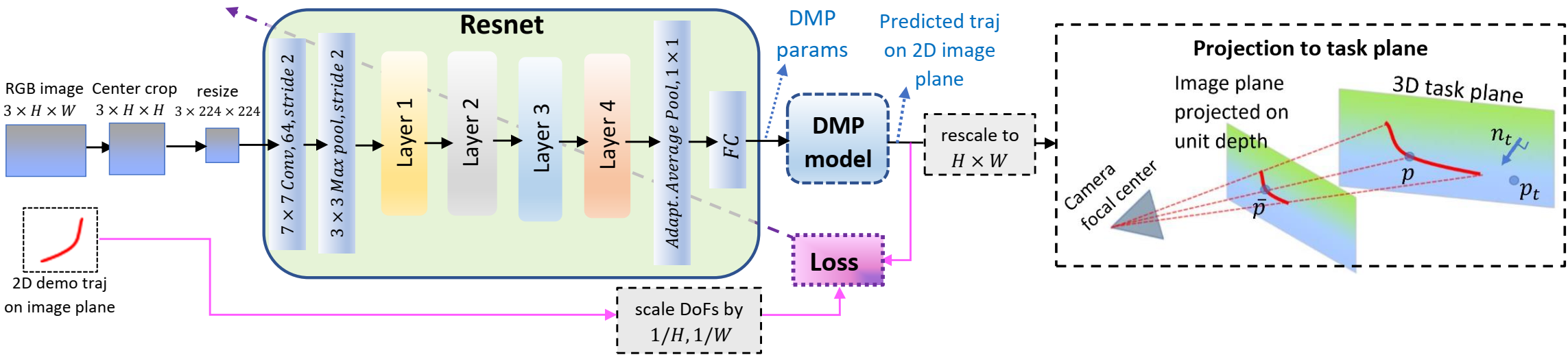}
    	\caption{Resnet-DMP architecture. Magenta arrows apply only during training.}
	\label{fig:resnet_dmp}
\end{figure*}

In this work we consider planar tasks, where the objective is to infer from a raw RGB image input of the scene the appropriate DMP parameters (weights that determine the motion pattern and initial and target position)  that will be used to generate a trajectory for solving the given planar task. 
To this end, we consider a demonstration dataset consisting of raw input RGB images along with the corresponding demonstrated trajectories on the image plane for executing the task, and design a network that will be trained on this dataset. Then, given a new input RGB image of the scene, the network should infer the appropriate parameters for the DMP, that generates the corresponding trajectory.
The desired time duration $T_f$ for executing the motion is preset in this work although it could in general be incorporated in the inferred DMP parameters.

The proposed architecture is shown in Fig. \ref{fig:resnet_dmp}, which we call Resnet-DMP. Input images are centered cropped by the smallest image dimension, resized to $224 \times 224$ and then given as input to a Resnet18\footnote{
The same architecture is applicable to other Resnet types as well. 
% \update{For details on the types and structure of Resnets see \ULurl{github.com/pytorch/vision/blob/main/torchvision/models/resnet.py}}
} that extracts from the RGB image the relevant latent features (after the adaptive average pooling block). The latter are fed to the FC (fully connected) layer block that infers the DMP parameters, which are used by the DMP model \eqref{eq:GMP_tf_sys} - \eqref{eq:GMP_y_x}, to generate the predicted trajectory on the image plane. 
For training, we calculate the smooth L$1$ loss between the predicted and the demonstrated trajectory, which is scaled by the initial dimensions of the image, hence the Resnet DMP learns normalized trajectories.
We opted for the smooth L$1$ over the L$2$ loss, because human demonstrations have an inherent variance, so for similar scenes, we want to penalize less trajectories that deviate more from the average behaviour.
For execution, the predicted (normalized) trajectory is first scaled by the dimensions of the image and then  projected
from the $2$D image plane to the $3$D task plane. This is done by using the camera's intrinsics (optical center and focal length) and its position in order to correspond each trajectory point 
% $\vect{p}_1 \in \mathbb{R}^2$ 
from the image plane to a point on the known $3$D task plane. This point is the intersection of the task plane with the ray from the camera focal center, passing through a trajectory point on the image projected to $3$D with unit depth (see Fig. \ref{fig:resnet_dmp}). Notice that the task plane may not in general be parallel to the image plane. Algorithm \ref{alg:image_to_task_proj} details the procedure.

% The latter is the intersection of the
% ray from the camera passing from the trajectory point on the
% image with the task plane. Notice that the task plane may
% not in general be parallel to the image plane. Algorithm 1
% details the procedure
 
% % $\bvect{p}$,
% using a unit depth, and then finding the desired point $\vect{p}$ from the intersection between the ray from the camera's position $\vect{p}_{c}$ along the vector $\bvect{p} - \vect{p}_c$, i.e. $\vect{p} = \vect{p}_c + \lambda(\bvect{p} - \vect{p}_c)$ with $\lambda \ge 0$, and the task plane $\vect{n}_t^T(\vect{p} - \vect{p}_t)=0$ with $\vect{n}_t$ being the task plane normal and $\vect{p}_t$ any position on the plane.}},

% using Algorithm . \cc{Or maybe remove the algorithm altogether and keep only footnote $3$? Also, maybe add $\vect{p}$, $\bvect{p}$, $\vect{p}_c$ in projection image in Fig. 1?}

% \cc{Stress a bit more this about the projection. One can be easily mistaken into thinking that this is the classical image to $3$D projection, which is not.}

% \cc{Explain why we use this projection and augment the data with perspective transformations, instead of employing heightmap.}

In \cite{DMP_NN_Ude_2020}, which uses the classical \cite{Ijspeert2013}, and in \cite{Arc_DMP_NN_Ude_2021}, which uses a modified DMP formulation \cite{ArcLen_DMP},
% where the classical DMP formulation is used, 
integration of the DMP dynamical equations is required to obtain the predicted DMP trajectory. This results in additional computations (numerical integrations) to obtain the partial derivatives of the loss w.r.t. the neural network's weights during back-propagation. 
However, using the novel DMP formulation \cite{Antosidi_Rev_DMP}, these additional computations can be circumvented, by observing that the novel DMP's dynamical equations \eqref{eq:GMP_tf_sys} - \eqref{eq:GMP_x_dot} result in $\vect{y}(t) = \vect{y}_{\phVar}(t) \ \forall t$ in the absence of any coupling terms. 
Therefore during training, we can use directly \eqref{eq:GMP_y_x} to obtain the predicted trajectory, without numerical integration.
Moreover, instead of predicting directly the DMP parameters $\vect{W}, \vect{g}, \vect{y}_0$ and plugging them in \eqref{eq:GMP_y_x} to obtain the DMP trajectory, we instead have the FC layer in Fig. \ref{fig:resnet_dmp} predict some new weights $\vect{W}' \in \mathbb{R}^{n \times K}$ anchored between the initial and target position, i.e. we re-parameterize the DMP position as $\vect{y}_{\phVar} = \vect{W}' \vphi(\phVar)$, hence $\vect{y}_0 = \vect{W}' \vphi(0)$, $\vect{g} = \vect{W}' \vphi(1)$. This facilitates the optimization, leading to smaller errors between the predicted and demo trajectories, which can be attributed to the reduction of the non-linearity of $\vect{y}$ (and in turn the loss function) w.r.t. the predicted DMP parameters.
% notice that predicting separately the DMP parameters $\vect{W}, \vect{g}, \vect{y}_0$ and plugging them in \eqref{eq:GMP_y_x} to generate the DMP trajectory would incur further non-linearity in the loss function. Instead, to facilitate the optimization of the Resnet's weights by reducing non-linearities,   we parameterize the DMP position directly as $\vect{y}_{\phVar} = \vect{W}' \vphi(\phVar)$, where $\vect{W}' \in \mathbb{R}^{n \times K}$ are some new weights anchored between the initial and target position, i.e. $\vect{y}_0 = \vect{W}' \vphi(0)$, $\vect{g} = \vect{W}' \vphi(1)$. 
% Therefore the FC layer in Fig. \ref{fig:resnet_dmp} contains $n \times K$ units. 
During execution we simply set $\vect{W} = \vect{W}'$ and to generate the DMP trajectory we use \eqref{eq:GMP_tf_sys}, \eqref{eq:GMP_x_dot} (where one can in general include coupling terms depending on the task).

\RestyleAlgo{ruled} 
\SetKw{MyInput}{Inputs:}
\SetKw{MyOutput}{Output:}
\newcounter{myalgocounter}
\setcounter{myalgocounter}{1}
\newcommand\algoline[1]{\themyalgocounter. \addtocounter{myalgocounter}{1} #1 \\}
\newcommand\algocomment[1]{{\color{LimeGreen}{\footnotesize \# #1 }}}

\begin{algorithm}
\caption{Image to task plane projection } \label{alg:image_to_task_proj}
\MyInput{\\}
\myinput{- $x, y$ : {\small trajectory point on the image plane}}
\myinput{- $c_x, c_y$, $f_x, f_y$ : {\small camera’s optical center and focal length}}
\myinput{- $\vect{n}_t$, $\vect{p}_t$ : {\small task plane normal and any point on that plane expressed in the world frame}}
\myinput{- $\vect{p}_{wc}$, $\vect{R}_{wc}$ : {\small Cartesian position and rotation matrix of the camera w.r.t. the world frame}}
\MyOutput{\\}
\myinput{- $\vect{p}$ : {\small $3 \times 1$, projection of $(x, y)$ on the task plane, expressed in world frame}}

\algoline{$\bvect{p} = [(x – c_x) / f_x \ (y – c_y) / f_y \ 1]$ \algocomment{project on unit depth}}
\algoline{$\vect{n}_t = \vect{R}_{wc}^T \vect{n}_t$ \algocomment{Convert to camera frame}}
\algoline{$\vect{p}_t = \vect{R}_{wc}^T (\vect{p}_t - \vect{p}_{wc})$}
\algoline{$\lambda = (\vect{n}_t^T \vect{p}_t) / (\vect{n}_t^T \bvect{p})$ \algocomment{calculate the actual depth}}
\algoline{$\vect{p} = \lambda \bvect{p}$ \algocomment{scale to the actual depth}}
\algoline{$\vect{p} = \vect{R}_{wc}\vect{p} + \vect{p}_{wc}$ \algocomment{convert from camera to world}}
\end{algorithm}

% Notice that thanks to the adopted DMP formulation which was shown in \cite{Antosidi_Rev_DMP} to be mathematically equivalent to the original DMP \cite{Ijspeert2013}, no additional manual calculations are required during training for back-propagation in contrast to \cite{DMP_NN_Ude_2020, Arc_DMP_NN_Ude_2021}.

%%%%%%%%%%%%%%%%%%%%%%%%%%%%%%%%%%%%%%%%%%%%%%%%%%%%%%%%%%%%%%%%%%%%
%%%%%%%%%%%%%%%%%%%%%%%%%%%%%%%%%%%%%%%%%%%%%%%%%%%%%%%%%%%%%%%%%%%%

\section{Experimental validation} \label{sec:Experiments}

To test our method we examine two planar tasks. The first one concerns the problem of unveiling the stem of a grape-bunch, that is occluded by leaves. This task serves both as a validation of our method and as a comparison with \cite{Arc_DMP_NN_Ude_2021}.
Then, to further validate our method we explore a more complex task that involves grasping an object surrounded by clutter (other objects) on a flat surface (task plane),  by pushing aside the other objects to free up the space around the target object and grasp it. Details for each experimental scenario are provided in the ensuing subsections. A video with indicative experimental trials can be found at 
\noindent{{\small \ULurl{https://youtu.be/t2g59403N6c}}

\subsection{Experimental testing and comparison in stem unveiling} \label{sec:Experiments_a}

% To test our method we examine 
To validate our method and compare it against \cite{Arc_DMP_NN_Ude_2021} we consider
the problem of unveiling the stem of a grape-bunch, that is occluded by leaves. 
Such a procedure is required as a preliminary step in case of an occluded stem in order to enable  reaching and cutting of the stem of a grape-bunch during harvesting and it consists a part of the harvesting procedure that was developed within the BACCHUS project\footnote{
\ULurl{https://bacchus-project.eu}}. 
We treat this as a planar task, where the task plane is determined by the vineyard plane, which is assumed to be known. 

\subsubsection{Experimental Set-up} \label{sec:Problem}

Our experimental set-up consists of two ur5 manipulators, one responsible for the unveiling of the stem using a custom $3$D printed gripper attached on its wrist and the other with a realsense2 vision sensor on its wrist (in-hand camera) providing the  RGB image of the scene (with resolution $480 \times 640$), as shown in Fig.\ref{fig:exp_setup}. 
A mock-up vine set-up is used with up to three plastic grapes and multiple leaves. The unveiling is to be performed w.r.t. the vision sensor.
For scenes with multiple occluded grapebunches, we assume that the system targets the one that is closer to the center of the image.
The gripper has a fixed aperture, and during each unveiling its orientation is fixed so that the gripper's axis is parallel to the camera's z-axis and the gripper's fingers are normal to the direction vector between the initial and target position of the projected image plane (Fig. \ref{fig:resnet_dmp}).
% with unit depth.}
% projection to $3$D with unit depth of the initial and target position from the image plane.}

\begin{figure}[!ht]
	\centering
	\includegraphics[scale=0.21]{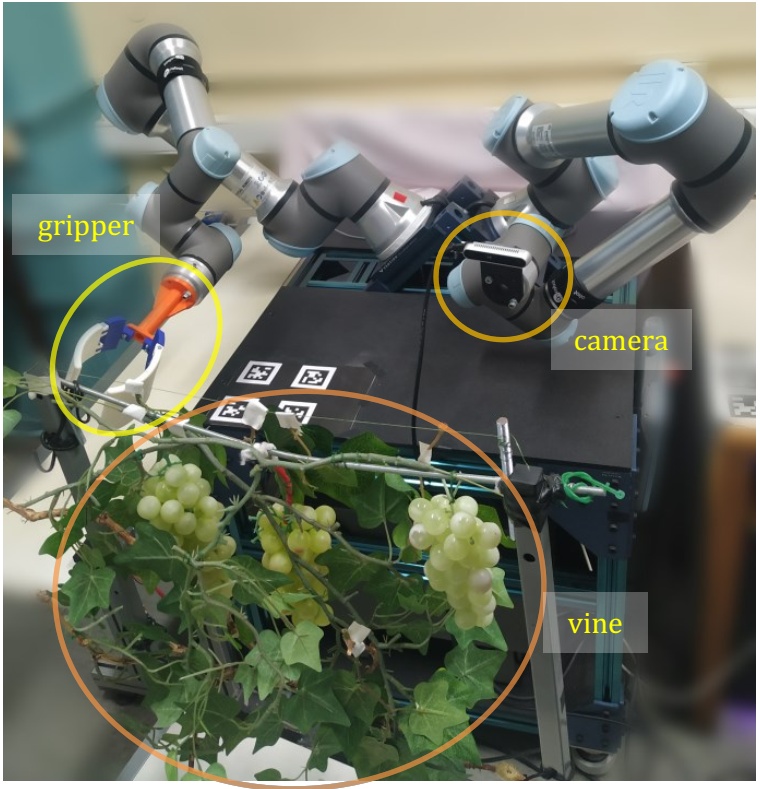}
    	\caption{Stem unveiling task setup.}
	\label{fig:exp_setup}
\end{figure}

\subsubsection{Collection of Demonstrations and Training} \label{sec:Demo_training}

\begin{figure}[!ht]
	\centering
	\includegraphics[scale=0.42]{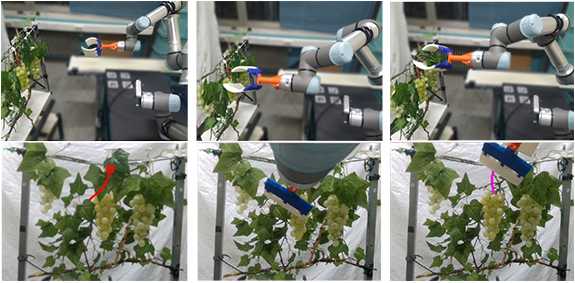}
    	\caption{Demonstration procedure. Top: scene side view. Bottom: In-hand camera view. The unveiling trajectory is drawn on the rgb image (shown with red on the bottom left image) and is then projected on the vine plane and executed by the robot. The unveiled stem is shown with magenta on the bottom right image.}
	\label{fig:sample_demo}
\end{figure}

\begin{figure}[!ht]
	\centering
	\includegraphics[scale=0.46]{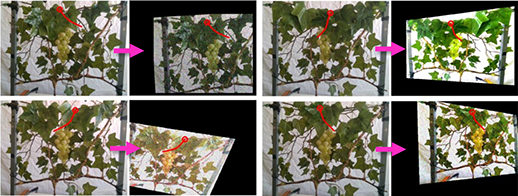}
        \caption{Data augmentation. The unveiling trajectory is shown with red. The magenta arrow shows the augmentation of each image.}
        % \includegraphics[scale=0.23]{images/augmented_samples.png}
    % \caption{Data augmentation. Left: initial image and unveiling trajectory with red. Right: Augmented image and corresponding unveiling trajectory.}
	\label{fig:augmented_samples}
\end{figure}

Demonstrations are collected by randomly rearranging the grape-bunches and leaves on the mock-up vine set up and for each scene having a human draw on the image the unveiling trajectory with a mouse. The resolution of the image ($480 \times 640$) was sufficient to ensure accuracy when drawing the trajectory. This trajectory is then projected from the image plane to the $3$D vine plane and executed by the robot as shown in Fig. \ref{fig:sample_demo}. If the stem of the vine is fully unveiled, the RGB image and the human demonstration are stored, otherwise the process is repeated.
All demo data samples are collected with the camera $0.68$ m from the vineyard plane and the camera's RGB image plane being parallel with the vine (task) plane, with a white background. During testing, other camera distances and viewpoints are used to assess the model's generalization capabilities in the more realistic cases that are expected during the entire harvesting procedure.

The basic idea is to learn from a reasonable amount of successful demonstrations, and then rely on artificial data augmentation to achieve better generalization performance.
Specifically, we collected $143$ samples (in about $3$ hours), split them randomly in train/dev/test sets with $93/25/25$ samples respectively, and then augmented each sample $100$ times by applying random $2$D translations, rotations, scaling, random horizontal flip and perspective transformations. In the RGB images we further apply color jitterring (random changes of brightness, contrast, saturation and hue) and add Gaussian noise. Samples of this augmentation process are visualized in Fig. \ref{fig:augmented_samples}.

The exact same augmented datasets were used to train end-to-end from scratch both the proposed Resnet-DMP model, as well as the VIMEDNet from \cite{Arc_DMP_NN_Ude_2021}. To make the comparison fair, we used the DMP formulation presented in \ref{sec:DMP_prelim}, with $K=10$ kernels for each DoF. Training was carried out for $150$ epochs with a batch size of $32$, using the Adam optimizer with initial learning rate $0.001$ that was halved every $40$ epochs. After training, to avoid over-fitting and achieve better generalization,
we kept the model from the epoch with the minimum dev-set loss.

% Initial evaluation test to assess the applicability and potential of our method were first carried out in simulations, which are deferred in Appendix A. In the following we describe the evaluation and testing performed on the actual setup.

\subsubsection{Evaluation and Experimental testing/comparison} \label{sec:Testing}

% The RMSE error for each dataset are reported in Table \ref{tab:real_results},
The training loss for each of the train/dev/test sets are plotted in Fig. \ref{fig:real_loss}. The epoch for the best model (w.r.t. the dev-set) is indicated with an arrow. 
The errors for the best ResnetDMP and VIMEDNet models are also reported in Table \ref{tab:real_results}, where we calculated the RMSE between each demo trajectory and the predicted one calculated over $50$ points, and computed the mean and standard deviation of this error across the samples of each dataset.
These results indicate that the Resnet-DMP fits accurately the train-set while also retaining smaller loss and RMSE errors compared to VIMEDNet. Moreover, VIMEDNet appears to show signs of over-fitting and the higher errors in the dev and test sets are indicative of unsatisfactory generalization to new scenes.
% While both models are able to fit quite accurately the train-set (the blue lines converge to zero), the VIMEDNet loss on the dev and test set exhibits a slightly increasing trend as the epochs progress, which is indicative of overfitting to the train-set and inadequate generalization to new data. In contrast, the Resnet-DMP manages to reduce the error in the dev and test sets, while still fitting accurately the train-set.}
% To further evaluate the training performance of each model we calculated the RMSE between each demo trajectory and the predicted one calculated over $50$ points, and computed the mean and standard deviation of this error across the samples of each dataset.
% The results are reported in Table \ref{tab:real_results}.
% % where it
% As expected, the RMSE error of VIMEDNet in the train set is higher than Resnet-DMP, since the best VIMENet model has higher train-loss compared to the best Resnet-DMP model, as can be seen from the intersection of the blue lines with the magenta vertical line in Fig. \ref{fig:real_loss}. The RMSE error for the dev and test sets are also higher for VIMEDNet,
% % It can be observed that despite both models having relatively close error in the train-set, the VIMEDNet exhibits much larger errors in the dev and test sets, 
% which is indicative of unsatisfactory generalization to new scenes. 
This can also be seen in the unveiling samples from the test set shown in Fig. \ref{fig:test_set_samples}.
Although VIMEDNet was successfully applied in \cite{Arc_DMP_NN_Ude_2021} for reproducing the path of a depicted digit, its generalization is not as good in our case.
This can be attributed to the fact that in \cite{Arc_DMP_NN_Ude_2021} a large artificially created demo training set was used with more concrete features relating strongly to a white path on a grayscale image, while in our task, the features that determine an appropriate unveiling trajectory are more subtle and fuzzy.
Despite, the relative small demo dataset in our case and the subtlety of the scene features, the more sophisiticated structure of the Resnet-DMP leads to much better generalization.

% two reasons: 1) In \cite{Arc_DMP_NN_Ude_2021} the task allowed for creating programmatically a very large train-set with thousands of artificial demonstrations. Having access to such a large train-set their model managed to generalize successfully. In our application though, and in many other practical scenarios, obtaining such a large dataset would be extremely time-consuming and impractical (note that $143$ demos took about $3$ hours). 2) The task in \cite{Arc_DMP_NN_Ude_2021} had more concrete features (relating strongly to a white path on a grayscale image), while in our task, the features that determine an appropriate unveiling trajectory are more subtle and fuzzy.
% Despite, the relative small demo dataset and the subtlety of the scene features, the more sophisiticated structure of the Resnet-DMP leads to much better generalization.}

\begin{figure}[!ht]
	\centering
	\includegraphics[scale=0.34]{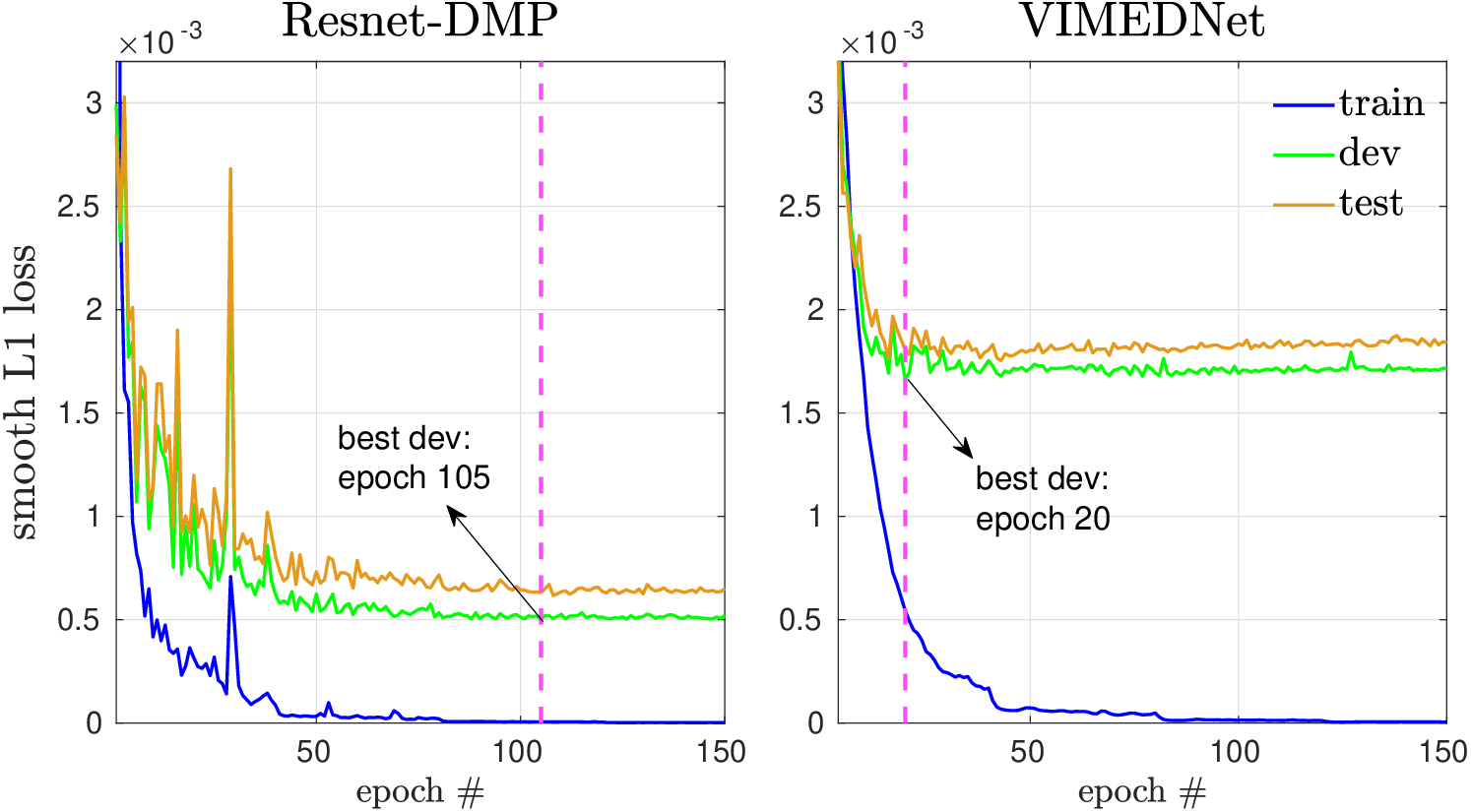}
    	\caption{Resnet-DMP vs VIMEDNet training loss. For each model, the network weights from the epoch with the lowest dev-set loss are kept (indicated with the dashed magenta line).}
	\label{fig:real_loss}
\end{figure}

We further carried out unveiling experiments on the mock-up vine setup, on $27$ new created scenes, where we also changed the camera view-point in each test, to make the task more challenging. 
The distance of the camera from the vineyard plane varied in $[0.5 \ 1]$ m.
% Remember that, the demonstrations were performed from a fixed view-point, perpendicular to the vine plane. Therefore, we rely solely on the data augmentation for the model to achieve generalization from different view-points.
To perform each unveiling, the robot with the gripper first reaches the initial point of the unveiling trajectory with a pre-planned trajectory,
% approached the initial point of the unveiling trajectory, with a depth offset of $4$ cm along the camera's depth axis, using a $5$th order polynomial in the Cartesian space, and then moved again with a $5$ order polynomial to reach the initial unveiling point. 
and subsequently follows the unveiling trajectory generated by the DMP with a fixed time duration of $4$ sec.
Both the Resnet-DMP and the VIMEDNet were tested on the same scenes, where the Resnet-DMP achieved $85\%$ unveiling success, while VIMEDNet only $52\%$. Some sample execution instances are shown in Fig. \ref{fig:real_test_exp}, where successful attempts are highlighted with green and failures with red image frames. Finally, we tested the Resnet-DMP in $17$ scenes with a busy background, with snapshots shown in Fig. \ref{fig:reaL_test_clutter}. Despite that the images from these scenes differ significantly from those used during training, the Resnet-DMP was still capable of producing reasonable unveiling trajectories, achieving $59\%$ success rate, which is promising. In particular, we can expect that collecting some additional demos from the real vineyard, we will achieve comparable success results to this work.

\begin{figure}[!h]
	\centering
	\includegraphics[scale=0.43]{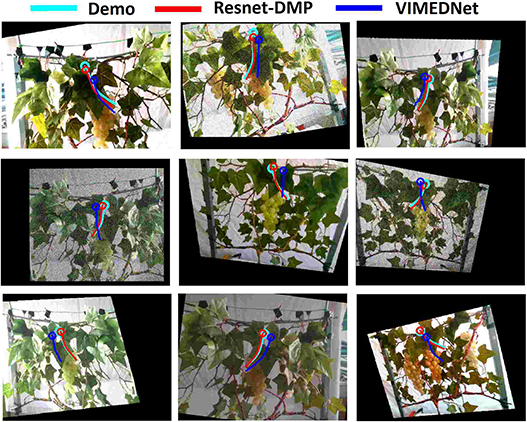}
    	\caption{Visualization of the predicted unveiling trajectory on instances of the test-set.}
	\label{fig:test_set_samples}
\end{figure}

\begin{table}[!h]
\centering
\caption{Resnet-DMP vs VIMEDNet. The first $3$ rows report the RMSE mean $\pm$ the standard deviation (in pixels) for each dataset, on $480 \times 640$ images. The last row reports the unveiling success rate on new scenes.}
\setlength{\tabcolsep}{0.55em} 
\begin{tabular}{c c c c }

\Xhline{1pt}
            &   & \textbf{Resnet-DMP} & \textbf{VIMEDNet} \\
\cline{2-4}
    \multirow{3}{*}{\rotatebox[origin=c]{90}{\parbox[c]{1cm}{\centering \textbf{RMSE}}}} & \textbf{train} & $1.77 \pm 0.49$  & $16.02 \pm 7.18$ \\
\cline{2-4}
    & \textbf{dev} & $14.88 \pm 11.77$  & $30.02 \pm 17.21$  \\
\cline{2-4}
    & \textbf{test} & $14.87 \pm 15.05$ & $32.12 \pm 15.16$ \\
\Xhline{1pt}
    \multirow{2}{*}{\rotatebox[origin=c]{90}{\parbox[c]{1cm}{\centering \textbf{Unveil success}}}} & {$27$ new scenes} & $85\%$ & $52\%$ \\
\cline{2-4}
   ~ & \makecell{$17$ new scenes \\ with busy background} {\footnotesize } & $59\%$ & - \\
\Xhline{1pt}
 \end{tabular}
\label{tab:real_results}
\end{table}

\begin{figure}[!ht]
	\centering
	\includegraphics[scale=0.42]{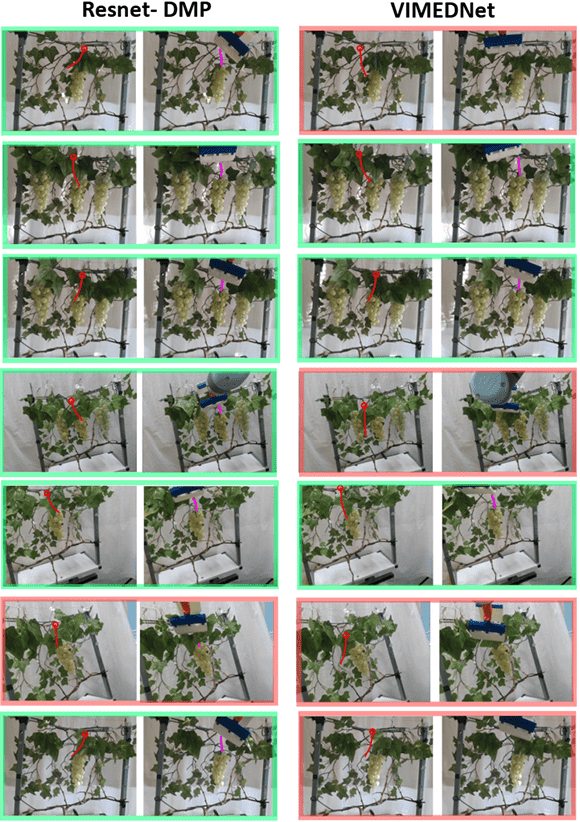}
    	\caption{Resnet-DMP vs VIMEDNet on real experiments. Successful unveilings are enclosed in green and failures in red frames. The part of the stem that is unveiled is highlighted with magenta.}
	\label{fig:real_test_exp}
\end{figure}

\begin{figure}[!h]
	\centering
	\includegraphics[scale=0.27]{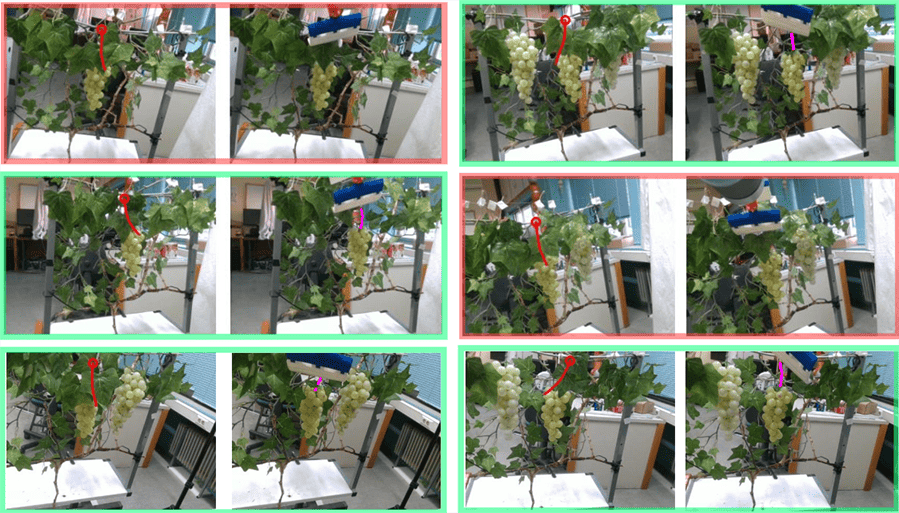}
    	\caption{Resnet-DMP test on scenes with busy background, not seen during training.}
	\label{fig:reaL_test_clutter}
\end{figure}

% \noindent{\small A video with indicative experimental unveiling trials can be found in} {\footnotesize \ULurl{dropbox.com/s/wqyqc5b9hgasz2d/video.mp4?dl=0}}

%%%%%%%%%%%%%%%%%%%%%%%%%%%%%%%%%%%%%%%%%%%%%%%%%%%%%%%%%%%%%%%%%%%%
%%%%%%%%%%%%%%%%%%%%%%%%%%%%%%%%%%%%%%%%%%%%%%%%%%%%%%%%%%%%%%%%%%%%

\subsection{Grasping an object on a cluttered surface by pushing aside the clutter} \label{sec5:Grasp_clutter}

To further evaluate our method, we consider the task of grasping an object, placed on a flat surface and surrounded by other objects (clutter), by pushing aside the clutter. 
The target object can be different in each scene. We utilize an RG2FT gripper, that rolls in its palm the target object and then grasps it by closing its fingers. The initial gripper aperture is chosen wide enough to ensure that all desired target objects can fit in this aperture. When the gripper closes its fingers to grasp a target object, it stops when the force measured between its fingertips exceed a predefined threshold. 
This approach can accommodate grasping a wide range of every-day objects with different geometries, unknown by the proposed method. Due to the presence of other objects, grasping the target object from the top may be difficult or even impossible, as the gripper's fingers may collide with the surrounding clutter. Another case where a top grasp is not possible is for instance when grasping an object from a shelf.
% Also, depending on the target object, grasping it from the top usually requires a more dexterous and precise grasp. 
For these reasons, we adopt a planar grasp, that is parallel to the surface where the objects are located. In order to grasp the target object amidst the clutter, the robot has to push aside the other objects that obstruct the grasping of the desired object and then grasp the target object. The push and grasp are performed in a single continuous motion.

\subsubsection{Experimental Set-up} \label{sec5:Grasp_clutter_Problem}

\begin{figure}[!ht]
	\centering
	\includegraphics[scale=0.25]{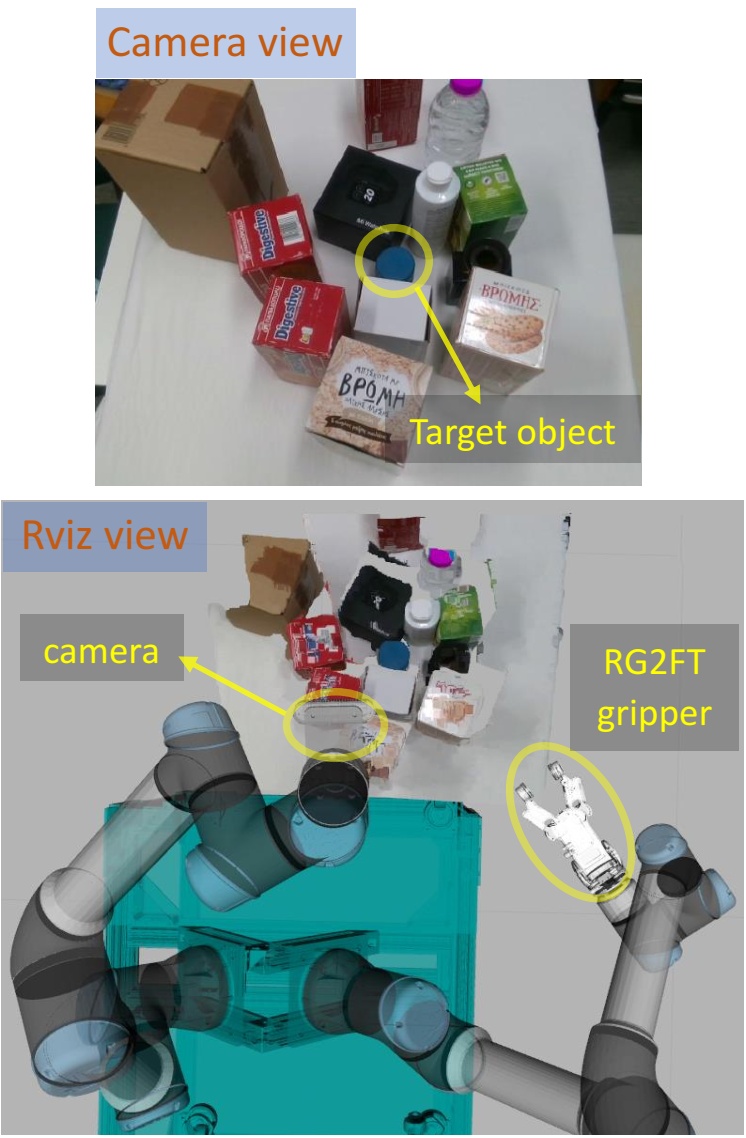}
    	\caption{Grasp object in clutter: task setup.}
	\label{fig:grasp_exp_setup}
\end{figure}

A visualization of the set-up is provided in Fig. \ref{fig:grasp_exp_setup}.
This set-up is similar to the one from Fig. \ref{fig:exp_setup}, with two ur5 manipulators, one responsible for pushing aside the clutter and grasping the target object and the second one equipped with a realsense2 vision sensor to provide the input image scene. The push-grasp operation is performed using the RG2FT gripper. 
Here the gripper's orientation plays a vital role and cannot be set heuristically, thus the Resnet-DMP will predict both the $xy$ planar Cartesian position as well as the rotation around the $z$ axis (yaw angle).
Both during training and testing, the pose of the ur5 with the camera is fixed, providing a top-down view of the task plane.
Moreover, since the target object is not always the same, apart from the RGB image, we also provide a bounding circle mask around the target object. The exact architecture is detailed in the following.

\subsubsection{Employed architecture} \label{sec5:Grasp_clutter_architecture}

\begin{figure}[!ht]
	\centering
        % \hspace*{-2cm}
	\includegraphics[scale=0.23]{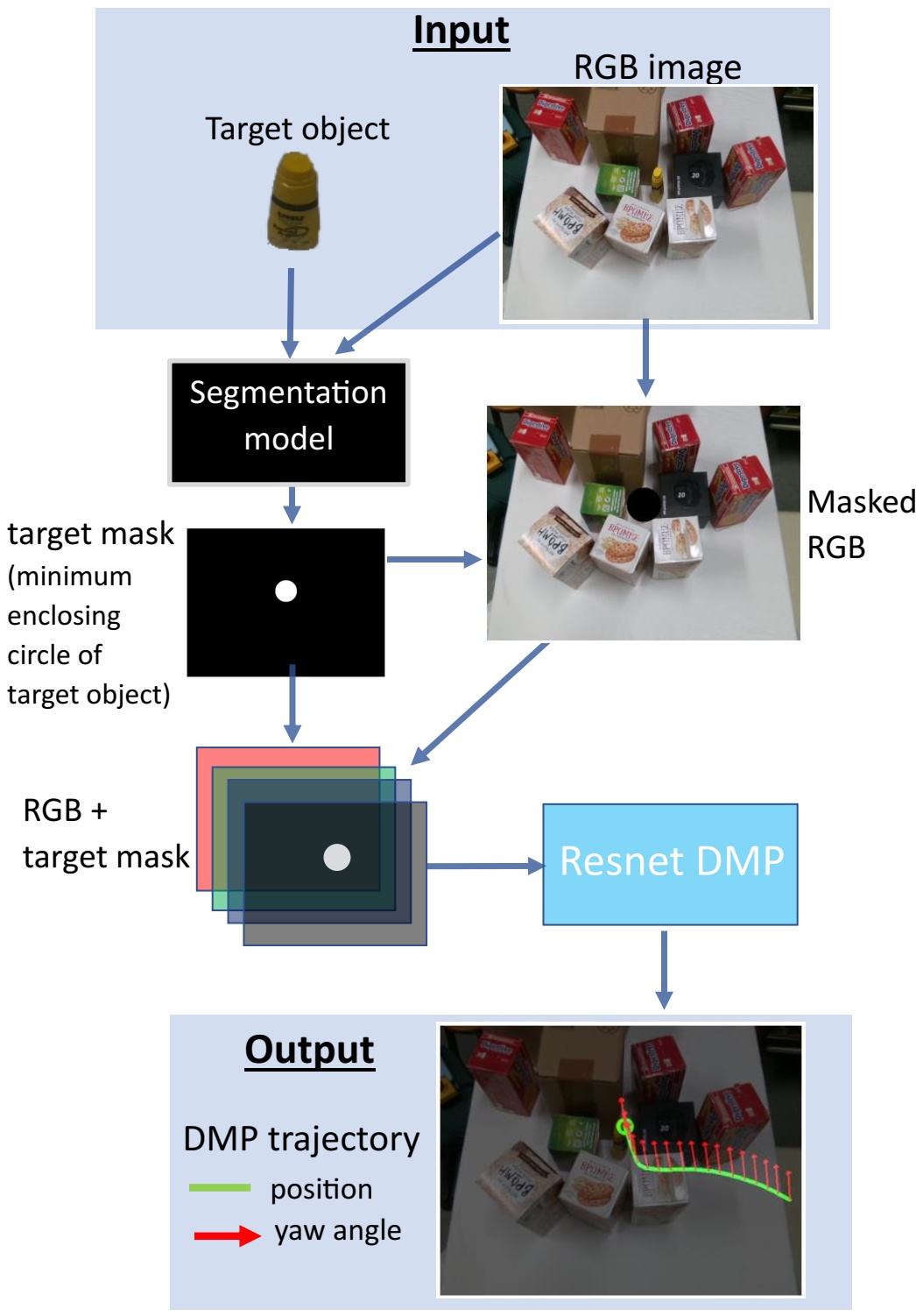}
    	\caption{Grasp object in clutter: architecture.}
	\label{fig:grasp_architecture}
\end{figure}

% We adapt the employed architecture-workflow for performing the grasping in clutter task. 

% First of all, here the gripper's orientation plays a vital role and cannot be set heuristically, thus the Resnet-DMP will predict both the $xy$ planar Cartesian position as well as the rotation around the $z$ axis (yaw angle). 
We employ the architecture depicted in Fig. \ref{fig:grasp_architecture}. The input consists of the target object and the RGB image of the scene. A semantic segmentation model is used to produce the mask of the target object from the RGB image and then a mask with the minimum enclosing circle of the target object is generated and used to mask-out the target object from the RGB image. This mask along with the masked RGB image are then provided as input to the Resnet-DMP (with its first convolutional layer modified to accept $4$ input channels) which then predicts the 2D oriented planar trajectory, i.e. $xy$ position and rotation around $z$ axis (yaw angle).
% , that is then back-projected to 3D using algorithm \ref{alg:image_to_task_proj}.

\subsubsection{Collection of Demonstrations and Training} \label{sec5:Grasp_clutter_Demo_training}

Demonstrations are provided through kinesthetic guidance, with the robot constrained so as to allow motion only along the $xy$ and/or around the $z$ axis. For all demonstrations the same target object was used. Since this target object is masked out before the image is given as input to the Resnet-DMP, the particular choice of the target object does not affect the model's inference during training or testing. 

\begin{figure}[!ht]
	\centering
	\includegraphics[scale=0.3]{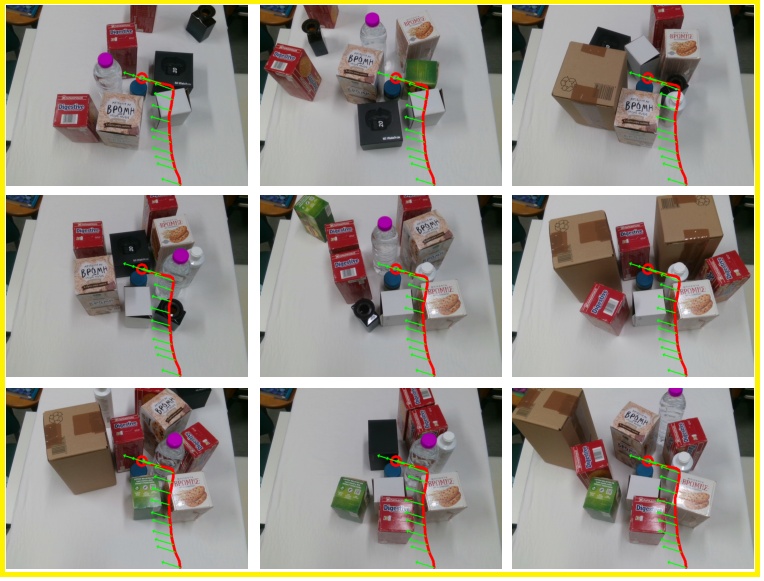}
    	\caption{Manual data augmentation.}
	\label{fig:grasp_manual_aug}
\end{figure}

The typical demonstration procedure would involve arranging some objects on the task plane, capturing the RGB image of the scene and then providing and recording the human demonstration and storing this particular training sample consisting of the image and the human recorded demo.
Instead of that, we follow a modified approach to the demo collection process, that allows us to collect much more training samples with less human demonstrations.
Specifically, for each human demonstration we capture additionally $N-1$ images in new scenes, created by inserting/removing objects or replacing objects with similar geometry in such a way that does not affect the initial demo.
% After $N \in [5 \ 20]$ rearrangements, the human performs the demonstration on the last scene. 
So with a single demonstration, a batch of $N \in [5 \ 20]$ training samples are stored. Examples of this procedure are shown in Fig. \ref{fig:grasp_manual_aug} for a particular demonstration, i.e. the same demonstration applies to multiply captured image scenes. 
% Of course, the larger the rearrangements are, the riskier it is to store also a training sample that could probably be inaccurate. However, the merit of having much more demonstrations outweighs the downsides of some potentially inaccurate training samples, whose effect should also diminish as more training data are collected. 

\begin{figure}[!ht]
	\centering
	\includegraphics[scale=0.32]{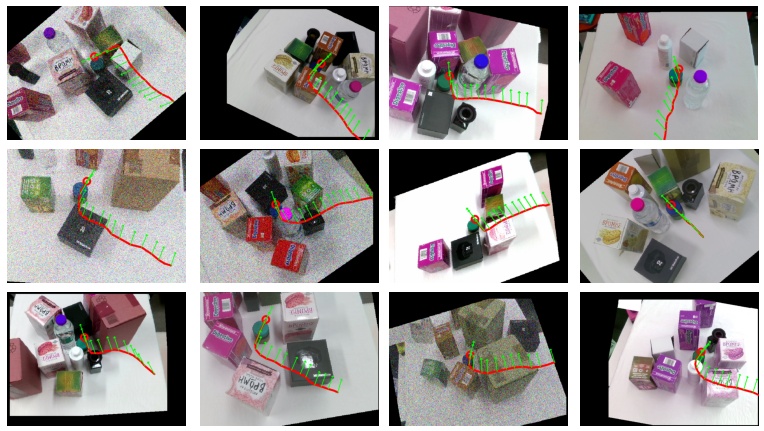}
    	\caption{Artificial data augmentation.}
	\label{fig:grasp_data_augmentation}
\end{figure}

% \begin{figure}[!ht]
% 	\centering
% 	\includegraphics[scale=0.58]{figures/grasp_train_loss.eps}
%     	\caption{Grasp object in clutter: Training loss.}
% 	\label{fig:grasp_train_loss}
% \end{figure}

% \begin{table}[!h]
% \centering
% \caption{Resnet-DMP RMSE mean $\pm$ the standard deviation for each dataset, on $480 \times 640$ images is reported.}
% \setlength{\tabcolsep}{0.55em} 
% \begin{tabular}{ c c c }

% \Xhline{1pt}
%                & \textbf{pos} (pxl) & \textbf{yaw-angle} (degrees) \\
% \Xhline{1pt}
%     \textbf{train} & $9.91 \pm 4.27$  & $1.70 \pm 1.17$ \\
% \Xhline{0.5pt}
%     \textbf{dev} & $28.07 \pm 27.52$  & $7.24 \pm 10.32$  \\
% \Xhline{0.5pt}
%     \textbf{test} & $29.66 \pm 41.47$ & $6.57 \pm 9.91$ \\
% \Xhline{1pt}
%  \end{tabular}
% \label{tab:grasp_sim_results}
% \end{table}

% \begin{figure}[!h]
% 	\centering
% 	\includegraphics[scale=0.5]{images/grasp_test_set_samples.png}
%     	\caption{Visualization of the predicted push-grasp trajectory on instances of the test-set.}
% 	\label{fig:grasp_test_set_samples}
% \end{figure}

Following the above procedure, a total of $305$ training samples were stored, performing only $16$ distinct demonstrations. These samples were randomly split in train/dev/test sets by $230/50/35$ and each sample in each set was further augmented artificially $100$ times using color jittering, injecting Gaussian noise and performing random translations, rotations, scalings and perspective transformations, as shown in Fig. \ref{fig:grasp_data_augmentation}. 
Training was carried out for $100$ epochs using the same hyper-parameters as in section \ref{sec:Demo_training}. Since in this case the loss function involves position and orientation expressed by the yaw-angle, prior to applying the smooth $L1$ loss, we divide the orientation by $2\pi$, so that it is in the range $[0 \ 1]$ like the position.
To avoid over-fitting and achieve better generalization, we kept the model from the epoch with the minimum dev-set loss.
% , which in this case was the model at epoch $27$, as can be seen in Fig. \ref{fig:grasp_train_loss}. The RMSE error of this model in each dataset is also reported in Table \ref{tab:grasp_sim_results}, while some predictions of the Resnet-DMP on instances of the test-set are shown in Fig. \ref{fig:grasp_test_set_samples}.

\subsubsection{Experimental testing} \label{sec5:Grasp_clutter_Testing}

We tested experimentally the proposed method in $80$ scenes with many objects (clutter) placed randomly on the task plane. For the target object we used $4$ different objects, choosing them as target with approximately equal frequency. A trial was considered successful if the target object was successfully grasped and retrieved. As the whole process is executed in open-loop, i.e. during execution the DMP does not receive any feedback regarding the target object or the scene in general, displacements of the surrounding objects may slightly perturb the initial pose of the target object. Therefore, during a specific trial, if the robot fails to grasp the object in the first attempt, we execute the DMP in reverse to retract the robot, and then the Resnet-DMP makes a new prediction on the new modified scene which is subsequently executed by the robot. This approach of repetitive attempts is common in the literature in similar tasks. For instance in \cite{Sarantopoulos_2021}, linear motion primitives are employed to isolate a target object within a clutter, requiring however many repetitions, while using a rich motion pattern, like DMP here, can complete the task even with a single attempt.
A trial is considered unsuccessful if the target object falls over or it has moved out of the camera view. 

\begin{figure}[!h]
	\centering
	\includegraphics[scale=0.25]{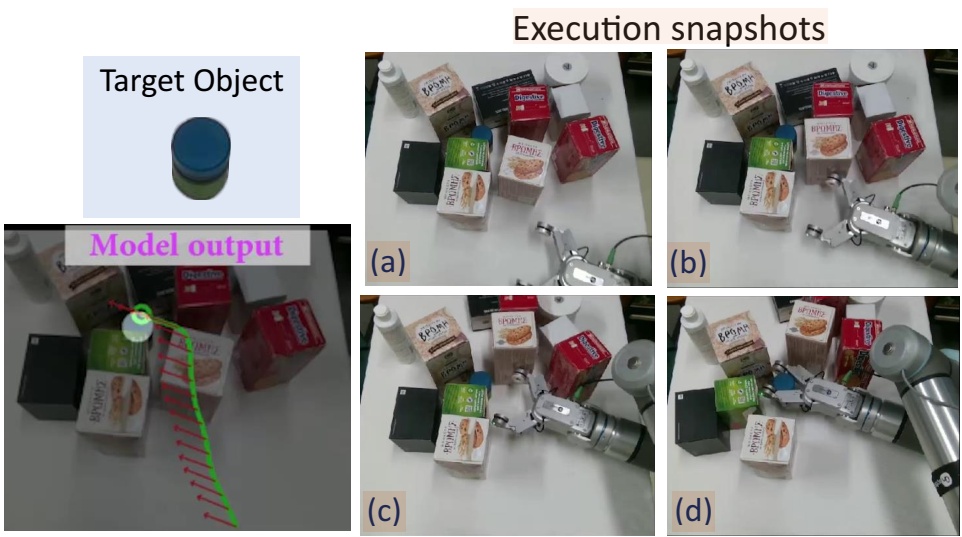}
    	\caption{Sample successful grasp trial 1.}
	\label{fig:grasp_ex1}
\end{figure}

\begin{figure}[!h]
	\centering
	\includegraphics[scale=0.25]{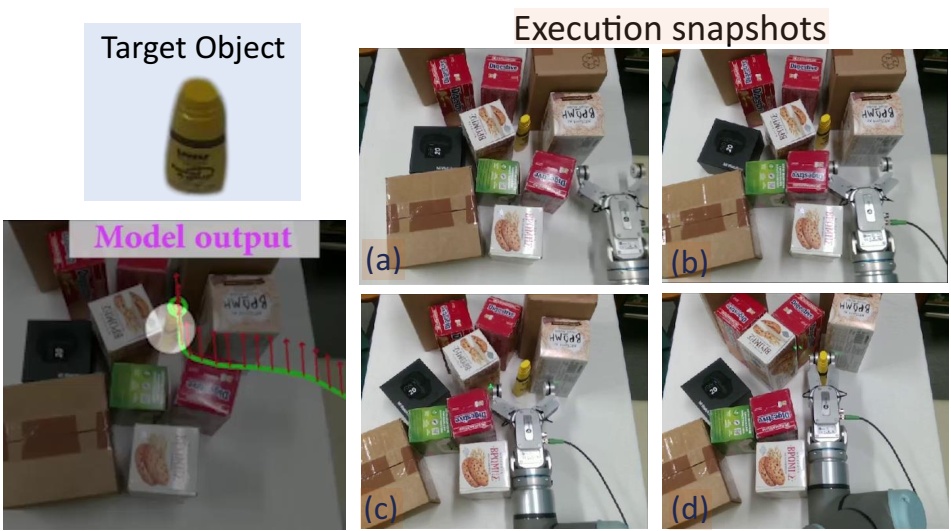}
        \caption{Sample successful grasp trial 2.}
	\label{fig:grasp_ex2}
\end{figure}

\begin{figure}[!h]
	\centering
	\includegraphics[scale=0.25]{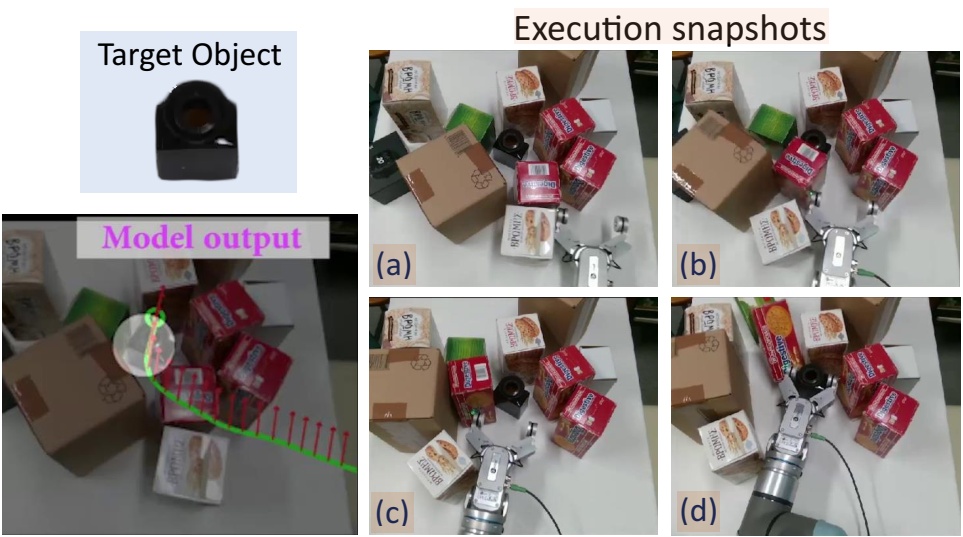}
        \caption{Sample successful grasp trial 3.}
	\label{fig:grasp_ex3}
\end{figure}

\begin{figure}[!h]
	\centering
	\includegraphics[scale=0.25]{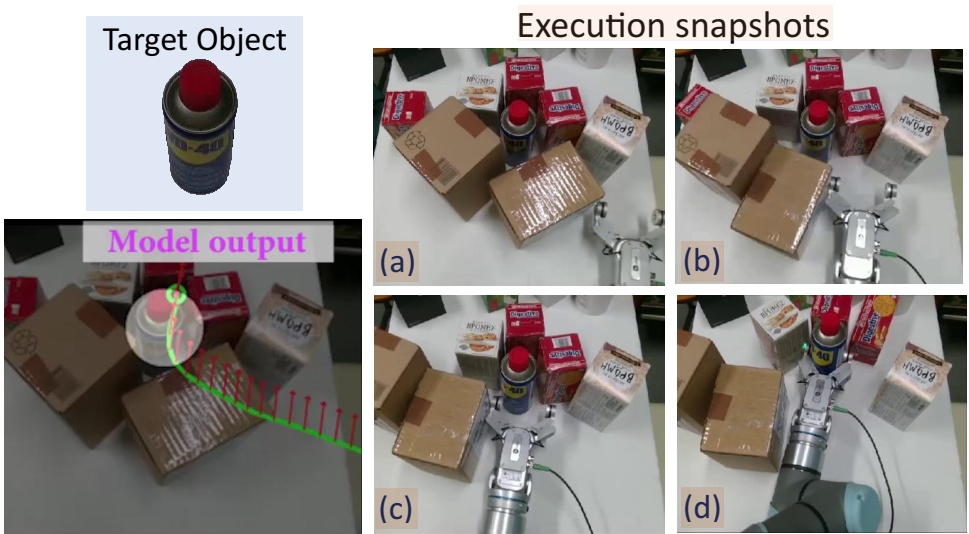}
        \caption{Sample successful grasp trial 4.}
	\label{fig:grasp_ex4}
\end{figure}

\begin{figure}[!h]
	\centering
	\includegraphics[scale=0.25]{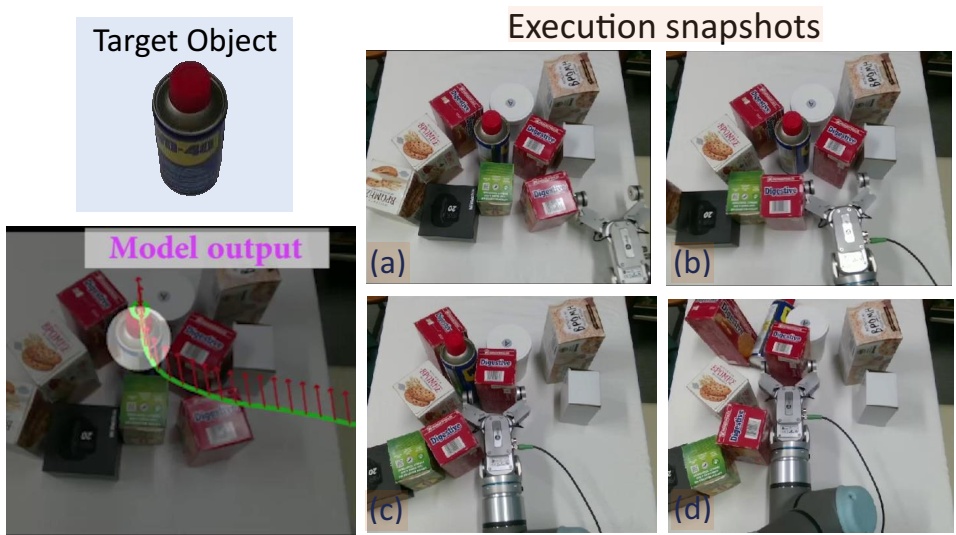}
        \caption{Sample unsuccessful grasp trial.}
	\label{fig:grasp_ex5}
\end{figure}

In total the model succeeded in $\bm{57/80}$ experiments, i.e. the \textbf{success rate} was $\bm{71.25 \%}$. Out of $57$ successful trials, $12$ required $2$ attempts while the rest were successful on the first attempt. Some successful executions are depicted in Fig. \ref{fig:grasp_ex1}-\ref{fig:grasp_ex4}, and an unsuccessful one in Fig. \ref{fig:grasp_ex5}. 
In each figure, on the left, the target object is shown and below it, the Resnet-DMP trajectory superimposed on the RGB image and the target object's circle mask. On the right, four snapshots of the execution are depicted.
% Further indicative experimental trials are provided in: \\ 
% {\footnotesize \ULurl{dropbox.com/s/34fw9d09tzvpqx2/video2.mp4?dl=0}}

%%%%%%%%%%%%%%%%%%%%%%%%%%%%%%%%%%%%%%%%%%%%%%%%%%%%%%%%%%%%%%%%%%%%
%%%%%%%%%%%%%%%%%%%%%%%%%%%%%%%%%%%%%%%%%%%%%%%%%%%%%%%%%%%%%%%%%%%%

\section{Conclusions} \label{sec:Conclusions}
In this work we employed a Resnet to infer the DMP parameters from raw RGB images for performing planar tasks. In this way, the proposed framework can encapsulate different motion patterns and infer automatically the appropriate one and the initial and target position for the generalization.
Our approach was validated and compared to another relevant SoA method in the task of unveiling the stem of grapebunch from occluding leaves. Our method was further experimentally validated for pushing other objects/clutter to grasp a target object on a flat surface. As future work we plan to extend our method to Cartesian tasks, using as input RGB-D images of the scene.

%%%%%%%%%%%%%%%%%%%%%%%%%%%%%%%%%%%%%%%%%%%%%%%%%%%%%%%%%%%%%%%%%%%%%%%

\bibliographystyle{IEEEtran}
\bibliography{main} 

\end{document}